# MEASURING NON-PROBABILISTIC UNCERTAINTY

*A cognitive, logical and computational assessment of known and unknown unknowns*

**Florian Ellsaesser**
**Frankfurt School of Finance & Management**

**Guido Fioretti**
**University of Bologna**

Contact: guido.fioretti@unibo.it

**Abstract**

There are two reasons why uncertainty may not be adequately described by Probability Theory. The first one is due to unique or nearly-unique events, that either never realized or occurred too seldom for frequencies to be reliably measured. The second one arises when one fears that something may happen, that one is not even able to figure out, e.g., if one asks: "Climate change, financial crises, pandemic, war, what next?"

In both cases, simple one-to-one cognitive maps between available alternatives and possible consequences eventually melt down. However, such destructions reflect into the changing narratives of business executives, employees and other stakeholders in specific, identifiable and differential ways. In particular, texts such as consultants' reports or letters to shareholders can be analysed in order to detect the impact of both sorts of uncertainty onto the causal relations that normally guide decision-making.

We propose structural measures of cognitive maps as a means to measure non-probabilistic uncertainty, eventually suggesting that automated text analysis can greatly augment the possibilities offered by these techniques. Prospective applications may concern actors ranging from statistical institutes to businesses as well as the general public.

**Introduction**

While most of the times uncertainty is nicely expressed by means of some probability distribution, there exist two sorts of problems where this is not the case. The first one occurs when probabilities must be measured on too small a sample, or no sample at all in the limit case of unique events. The second one occurs after some event has happened, that one had not been able to figure out. One may suspect that other surprising, currently unconceivable events may occur, a circumstance which generates uncertainty about the possibilities being envisaged rather than the relations between them. With possibly awkward, but certainly impressive expressions these two sorts of uncertainty have been labelled as due to "known unknowns" and "unknown unknowns," respectively (Rumsfeld, 2011; Feduzi & Runde, 2014; Faulkner *et al*., 2017). In both cases, neuro-psychological experiments involving observation of functional magnetic resonance imaging and electroencephalograms (Gluth *et al*., 2012, 2013, 2017) come along with research in logic and decision theory that, we shall suggest, lends itself to machine learning (ML) applications onto textual data.

In logic, the first case is exemplified by Ellsberg's conceptual experiment. Ellsberg (1961) asked to consider two urns, A and B. Urn A entails black and white balls in equal proportions, whereas all you know about urn B is that it entails black and white balls. Ellsberg remarked that, although the Principle of Sufficient Reason suggests to attach probability ½ to extract a white (or a black) ball from either urn, no-one would experience the same amount of uncertainty with B as with A.

From a purely conceptual point of view, two solutions exist for this conundrum, which correspond to the frequentist and the subjectivist interpretation of probability, respectively. According to the frequentist interpretation case A is equivalent to tossing a coin infinitely many times, whereas B amounts to expressing oneself on a sample of size zero. Thus, sample size marks the difference between A and B.

By contrast, within a subjectivist interpretation either this is not an issue (De Finetti, 1931) or, in more recent versions, sub-additive probabilities would be called to rescue (Gilboa, 1987; Schmeidler, 1989). According to this extension of the basic theory, probabilities are allowed not to sum up to unity if information is less than perfect. Thus, in the case of Ellsberg's paradox one would assign ½ to the probability of either extracting a black or a white ball from urn A, but zero probability to extract either a white or a black ball from B. Taking expectations would suggest that decision-makers prefer urn A, which conforms to common sense.

However, both the frequentist and the subjectivist solution show their limits if one is confronted with urns of type B only. If all options are such that little or no experience is available, measuring probabilities on nearly-zero samples or attaching very low sub-additive probabilities to all consequences is of little help. In economics, this has been recognized to be the case of uninsurable risk (Knight, 1921), where possibilities are known, but probabilities are unknown. It's the case of "known unknowns."

Neuro-psychological experiments show that while information is being gathered neuronal activity increases until it reaches a threshold, at which point a decision is finally made (Gluth *et al*., 2012). Correspondingly, decision theory theorizes of *satisficing behaviour* as making a decision as soon as a certain aspiration level is reached (March and Simon, 1958)[1] or, with greater emphasis on the need to accommodate conflicting options, *choice deferral* (Tversky and Shafir, 1992). In general, the received wisdom is that – as it is the case with uninsurable risk (Knight, 1921) – unless a sufficient amount of information has been gathered no decision can be made.

By contrast, we shall submit that even in this case decision-makers can analyse the structure of their mental representations of possibilities, looking for areas where alternative courses of action do not fan out into widely different consequences. In particular, we shall document this practice in the context of Scenario Planning and, most importantly, we shall point to the existence of geometrical representations of the intricacy of hypergraphs that can be used to assess this sort of uncertainty.

The second case has a more profound origin and involves more destructive consequences. The corresponding sort of uncertainty, which has been eventually qualified as "Keynesian," "fundamental," "true," "epistemic," "ontological" or "radical" uncertainty[2] (Runde, 1990; Davidson, 1991; Dunn, 2001; Dequech, 2004; Lane & Maxfield, 2005; Kay & King, 2020) arises when decision-makers fear that something may happen, that they are not even able to figure out. This sort of uncertainty is likely to be there if something that had not been conceived actually materialized, turning an "unknown unknown" – a possibility whose very existence is unknown (Rumsfeld, 2011; Feduzi & Runde, 2014; Faulkner *et al*., 2017) – into a so-called "black swan."

---

[1] According to Gluth *et al*.'s (2012) experiments, the threshold decreases with time. By contrast, March and Simon (1958) had theorized that the aspiration level would slowly increase once satisfaction is reached. These two statements could be possibly reconciled by observing that March and Simon had in mind real-world situations where decision-makers could search for better alternatives, a circumstance that in Gluth *et al*. (2012) was foreclosed by design.

[2] While Davidson (1991) contrasts "epistemic" to "ontological" uncertainty, taking the latter as a synonymous of probabilistic uncertainty, Lane and Maxfield (2005) employ the term "ontological uncertainty" in pretty much the same sense as Davidson's "epistemic" uncertainty.

Just like measures of sample size or sub-additive probabilities can formally extend utility maximization to include the case of imperfect information, incomplete preferences are eventually called in in order to deal with unknown unknowns (Eliaz & Ok, 2006; Ok, Ortoleva & Riella, 2012; Galaabaatar & Karni, 2013). We do not question the technical perfection of this solution but we stress that, just as it happened with sub-additive probabilities, technical perfection does not imply usefulness. With sub-additive probabilities and incomplete preferences utility maximization can be formally extended to encompass exotic types of uncertainty, but if those sub-additive probabilities are all close to zero and if no preference exists because alternatives cannot be formulated, expected utility maximization reduces to a hollow shell.

We rather propose an alternative route out of the observation that no uncertainty is there insofar as the unknown remains unknown. Once again, neuro-psychological experiments provide substantial insights.

According to Gluth *et al*. (2013, 2017), when subjects were faced with conflicting evidence neurons emitted both excitatory and inhibitory signals that made overall activity oscillate. These oscillations corresponded to decision being postponed, a state of mind which they labelled *deciding not to decide* (Gluth *et al*., 2013) in order to stress that it arose out of complex neuronal activity rather than rest.

In the above experiments, conflicting evidence was the origin of the inability to make a decision. We submit that the second sort of non-probabilistic uncertainty arises when something that had not been imagined – an unknown unknown – suddenly enters the set of possibilities envisaged by a decision-maker and upsets their network of causal relations.

Furthermore, we suggest that this sort of non-probabilistic uncertainty can be observed on mental representations of possibility sets. More specifically, we submit that occasional disruption of the graph of causal relations that link alternatives to consequences can reveal its emergence. In particular, we shall illustrate this sort of disruption in the self-representation of the BioTech industry with respect to "Big Pharmas."

Cognitive maps are essential in order to define measures for both sorts of non-probabilistic uncertainty. In their simplest version (Axelrod, 1976; Sigismund Huff, 1990; Sigismund Huff, Huff & Barr, 2000; Sigismund Huff & Jenkins, 2002), cognitive maps are network representations of world views whose nodes are concepts linked to one another by causal relations. Cognitive maps are traditionally obtained by analysing texts or recorded speeches, such as letters to shareholders, technical reports, or interviews, but we shall point also to the possibility of automatically extracting them by means of algorithms that can be potentially applied to very large amounts of data.

We develop our arguments in the ensuing two sections. In the first one, we propose structural measures that can be applied to causal relations of given alternatives and consequences when too little empirical evidence is available to measure probabilities (known unknowns). In the second one, we propose structural measures that can be applied when a novel, unexpected "unknown unknown" appears. In the first case, we point to Scenario Planning as a decision-making tool to which, in certain circumstances, our techniques could be applied. In the second case we point to disruptions of cognitive maps. The ensuing third section sketches machine learning algorithms that are able to speed up measurement by several orders of magnitude. A final section concludes with a general assessment of the prospects of our measurement techniques.

**The Complexity of Scenarios**

Scenario Planning emerged among business strategists as a procedure to become aware of available options through extensive discussion and intentional search for non-obvious possibilities that may upset the received wisdom (Schoemaker, 1995; Van der Heijden, 2000; Chermack, 2004; Roxburgh, 2009; Ramírez, Österman & Grönquist, 2013; Erdmann, Sichel & Yeung, 2015). The outcome of this exercise is a set of scenarios that have the purpose of preparing strategists for non-trivial future contingencies.

Scenarios constitute a network of concepts linked to one another by causal relations. Indeed, the network representation of scenarios is nothing but their authors' cognitive map (Goodier *et al.*, 2010; Amer, Jetter & Daim, 2011; Jetter & Schweinfort, 2011; Alipour *et al.*, 2017).[3] Although in many instances probabilities can be attached to the causal relations that lead to alternative scenarios and the scenario exercise becomes indistinguishable from taking expectations, Scenario Planning is most valuable when probabilities are unknown (Wilson, 2000; Goodwin & Wright, 2001; Wright & Goodwin, 2009; Ramirez & Selin, 2014). In this case, Scenario Planning is valuable insofar as it allows an analysis of the structure of causal relations that link known possibilities to one another, whose probabilities are unknown. We are, in other words, in the case of "known unknowns."

---
[3] These Authors generally adopt a probabilistic interpretation of scenarios. However, the connection they make between scenarios and cognitive maps remains equally valid.

For instance, the scenarios described in Figure (1) illustrate a possible cognitive map of the consequences of the Jan 2016 UN decision to lift sanctions on Iran in terms of oil exports (loosely inspired by Alipour *et al*., 2017). This cognitive map does not simply take account of the presumably larger offer of oil on the world market with the ensuing price dynamics, but also less obvious factors such as the availability of shale oil as well as renewable energy sources. Due to these factors, a non-obvious Scenario 2 appears along with the rather obvious Scenario 1. According to Scenario 2, in spite of lifting sanctions both production and revenues will stagnate, which is just the opposite of what one would expect.

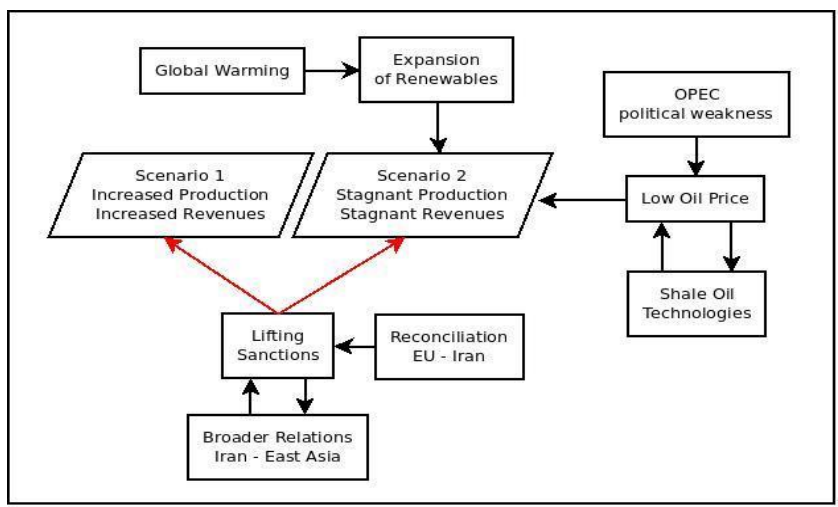

Figure 1. One and the same action - lifting sanctions on Iran - may lead to quite different outcomes depending on many other factors, such as growing availability of renewable energy sources or shale oil. One-to-many causal relations are highlighted in red. Loosely inspired by Alipour *et al*. (2017).

Notably, the authors of the scenarios illustrated in Figure (1) could not attach any probability to the causal relations that would yield either Scenario 1 or Scenario 2. Novel technologies and unprecedented political alliances made it impossible to provide a numerical estimate more reliable than a personal guess. Just like Ellsberg's urn B (Ellsberg, 1961), also in this case it all depends on the lack of a sufficiently large sample.

However, in spite of all uncertainty on probability values, Scenario Planning is far from useless. Indeed, it is even more useful than in the case probabilities could be attached to it, for without probabilities Scenario Planning is a tool for eliciting possibilities out of multiple points of view (Stirling, 2010). For instance, in the above example becoming aware of the non-obvious Scenario 2 was the true goal of the scenario exercise.

From a structural point of view, the interesting portion of Figure (1) is the one-to-many relation from "Lifting sanctions" to either Scenario 1 or Scenario 2. We maintain that it makes sense to analyse the structure of causal relations in exercises such as Scenario Planning in order to assess the uncertainty generated by such one-to-many causal relations.

In order to keep matters tractable let us assume that cognitive maps are made of *Evoked Alternatives* (EA), *Perceived Consequences* (PC), and causal relations from EA to PC (March & Simon, 1958). For instance, the relevant portion of the cognitive map of Figure (1) is made of one evoked alternative (lifting sanctions) and two perceived consequences (increasing or stagnant oil production, respectively).

Figure (2) illustrates three arrangements of causal relations between evoked alternatives and perceived consequences. On the left (a), all causal relations are one-to-one. This is the simple world where one perceives exactly which consequence follows from each of the evoked alternatives. In the middle (b) is the extremely complex world where one perceives that any consequence can follow from each of the evoked alternatives. On the right (c), is the somehow intermediate situation where, in spite of several one-to-many relations, one knows that certain evoked alternatives can only lead to a subset of the perceived consequences.

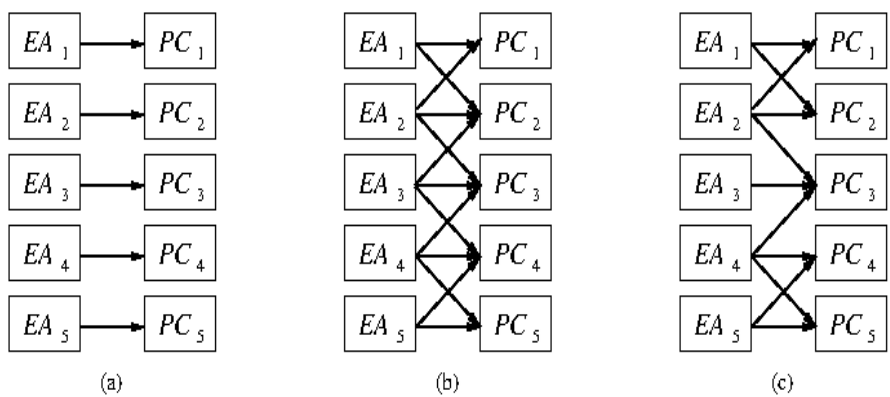

Figure 2. Three stylized cognitive maps linking evoked alternatives (EAs) to perceived consequences (PCs). In (a), a one-to-one mapping. In (b), the most confusing mapping where anything may happen. In (c), an intermediate case.

The case illustrated in (c) is the most interesting one, because the network of causal relations exhibits a structure. One feature of this structure is that although certain areas are tightly connected ($EA_1$ and $EA_2$ with $PC_1$ and $PC_2$; $EA_4$ and $EA_5$ with $PC_4$ and $PC_5$), these areas are more loosely connected with one another and therefore, even though it may be difficult to tell the difference between selecting $EA_1$ and $EA_2$, it is possible to state that either choice is different from either selecting $EA_4$ or $EA_5$. One other feature is that there are groups of perceived consequences that are unlikely to occur in isolation from one another (for instance $PC_2$ and $PC_3$ if one selects $EA_2$). Therefore, such sets of consequences constitute entities in themselves.

In algebraic topology such sets are called *simplices*, and their arrangement in a connected structure is a *simplicial family*, or *hypergraph* (Fioretti, 2023). Their structure can be analysed and subsumed by quantitative indicators based on *Q*-Analysis (see Appendix A). In a nutshell, each evoked alternatives $EA_i$ corresponds to a simplex whose vertices are the perceived consequences $PC_{ij}$ to which it is connected. Thus, the simplices that constitute the hypergraph have common faces whose dimensions depend on the number of perceived consequences that they have in common. The structure of the hypergraph is such that the areas of the decision problem where each evoked alternatives correspond to very many perceived consequences are represented by clusters of simplices connected along high-dimensional faces. These are the areas that generate high uncertainty. However, this uncertainty is eventually curbed if these areas are connected to one another by low-dimensional simplices as in Figure (2), case (c).

Decision problems where simplices share high-dimensional faces obtain the greatest benefits from *Q*-Analysis. In non-mathematical terms, this category includes all decision problems where different alternatives imply partially similar consequences, such as diagnoses of rare diseases whose symptoms appear in variable clusters of indicators and partially overlap with those of other diseases (Rucco *et al*., 2015), or alternative environmental policies that, through complex feed-backs in the ecosystem, may yield partially overlapping negative consequences (Eder *et al*., 1997; Forrester *et al*., 2015). Note that in the case of rare diseases, as well as with policies that have a long-lasting impact on the environment, no reliable probability is available.

Several measures can be defined on a hypergraph. In particular, it is possible to define suitable complexity measures (Casti, 1989; Fioretti, 2001). For instance, one such measure ascribes zero complexity to case (a), maximum complexity to case (b) and intermediate complexity to case (c) (see Appendix A). In a modelling exercise, boundedly rational agents could be assumed not to make any decision if complexity exceeds a threshold.

**When Unexpected, Novel, Destructive Possibilities Materialize**

Let us now focus on the reason why human beings may decide not to decide, namely when they become aware that some destructive possibility may materialise, that is not among those they are currently able to envisage. Such sentiments make perfect sense once unexpected and destructive events have been experienced, suggesting that unthinkable novelties may appear in a world that is governed by unknown laws, if at all. We submit that a series of cognitive maps taken before, at the time of, and after one such disruptive event allows us to observe such states of mind.

We illustrate our point with a series of cognitive maps in the BioTech industry extracted from the technical reports of *Ernst & Young*, a dedicated consultant (James, 1996; See also Appendix B). Among BioTech companies, a disruptive possibility materialized in 1990. Up to 1989 most of them were convinced that they would grow up to become able to produce and market their own drugs. For the time being they had to stipulate strategic alliances with pharmaceutical companies, but this was considered a temporary arrangement. In reality, their long-term goal was becoming "Big Pharmas." By contrast, pharmaceutical companies were entering strategic alliances with biotech companies in order to acquire their technology. Their long-term goal was generating biotech-based new drugs in-house. In 1990, biotech companies suddenly realized that many contracts that they had signed entailed "poison pills" designed to squeeze their knowledge and profits (James, 1996).

Figures (3) and (4) illustrate a portion of the biotech companies' cognitive maps in 1989 and 1990, respectively (James, 1996). Note that in 1990 the block *Poison Pills* entered the map, destroying previous linkages.

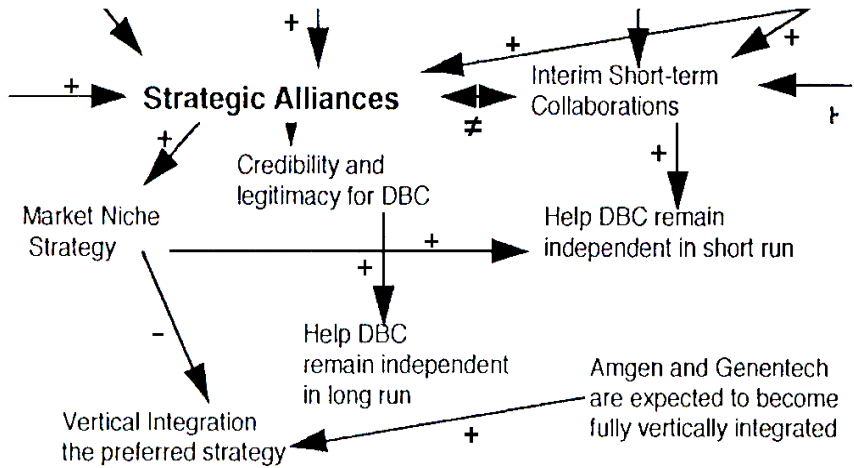

Figure 3. A portion of biotech companies' cognitive maps in 1989 (James, 1996). Incoming arrows stem from parts of the map not shown in this figure.

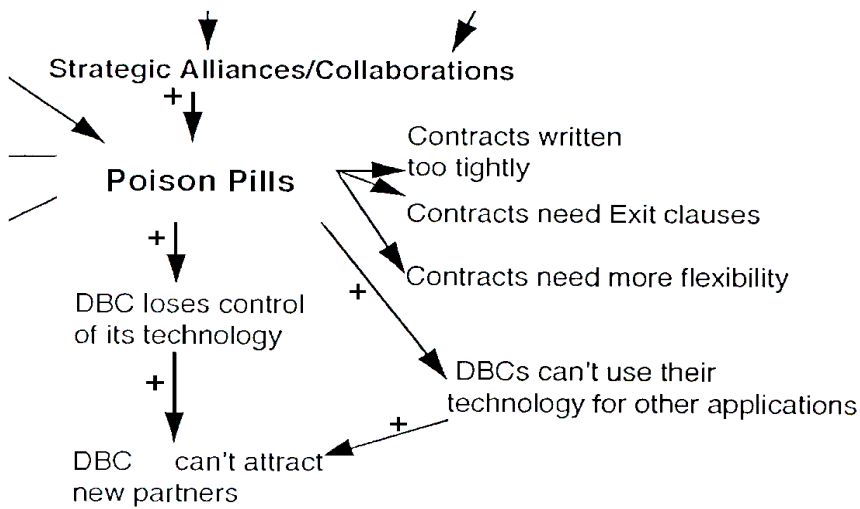

Figure 4. A portion of biotech companies' cognitive maps in 1990 (James, 1996). Incoming arrows stem from parts of the map not shown in this figure.

However, in 1991 biotech companies eventually understood the difficulties involved in drug production and distribution. Conversely, pharmaceutical companies realized that small and independent BioTech companies would guarantee a degree of exploration that in-house, hierarchically organized research could not attain (James, 1996). Thus, since 1991 the cognitive maps of biotech companies stabilized again, providing once again a reliable orientation to decision-making. Figure (5) illustrates a portion of the 1991 cognitive map.

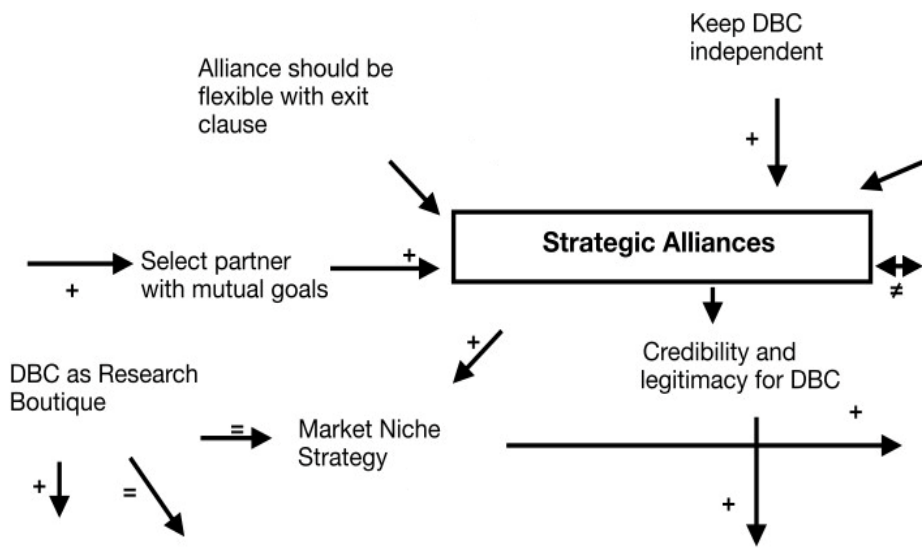

Figure 5. A portion of biotech companies' cognitive maps in 1991 (James, 1996). Incoming and outgoing arrows concern parts of the map not shown in this figure.

In the end, we have a series of eight cognitive maps 1986 to 1993, with one stable period 1986-1989, one other stable period 1991-1993, and one single disrupted cognitive map in between, in 1990. A few simple metrics can be explored in order to identify indicators that are able to single out the disrupted 1990 map from those of the stable periods that precede and follow it (Eden *et al*., 1992).

Figure (6) illustrates the number of concepts and its average (black), the number of linkages and its average (green) as well as the ratio between the number of linkages and the number of concepts (bottom line), which is remarkably constant over time. Notably, the number of linkages (green) exhibits a marked drop with respect to its average in 1990. The number of concepts (black) dropped, too, though possibly less sharply.

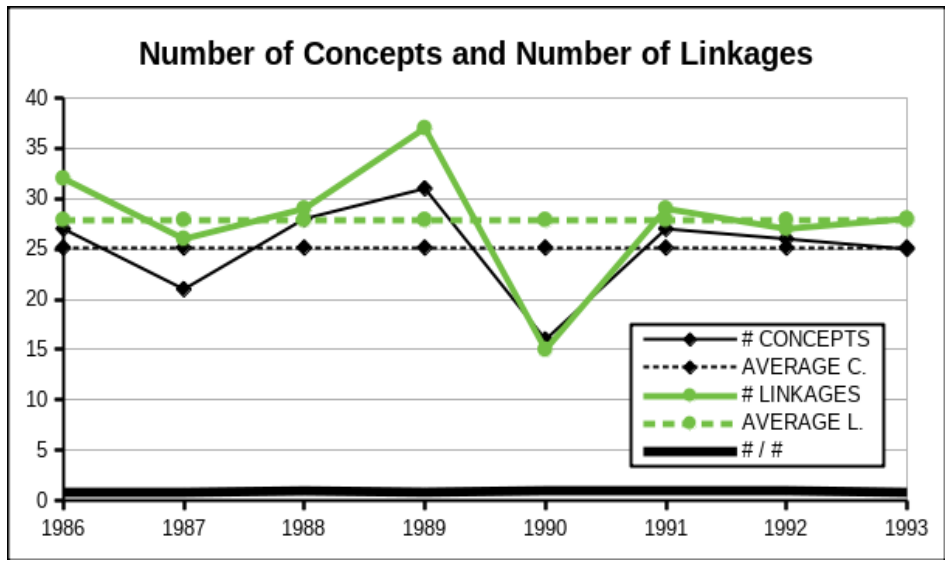

Figure 6. A few metrics of James (1996) series of cognitive maps. The number of concepts (solid black) compared to its average (dotted black), the number of linkages (solid green) compared to its average (dotted green), as well as their ratio (thick black). Notably, the ratio of these two metrics is extremely stable.

To our knowledge, the cognitive maps collected by James (1996) are unique in their ability to capture the magic moment of despair caused by "unknown unknowns" becoming true. The reason is that James (1996) extracted her cognitive maps from the technical reports of a consultant who was keen to express views that, possibly, no single BioTech company had ever confessed. To our knowledge, no other document has ever revealed a comparable richness of details, doubts and surprise.

Lacking other empirical examples, the proposed metrics – number of concepts, number of linkages – should be understood as a tentative, though sensible starting point. It seems sensible to assume, however, that collapsing cognitive maps display some sort of macroscopic structural change.

**Machine Learning**

The methods illustrated in the previous sections were based on eliciting cognitive maps out of text, which is a lengthy, painstaking activity. However, we submit that Machine Learning (ML) can open up novel possibilities in this respect.

ML is a rapidly expanding field that allows processing huge numbers of documents available on the Internet and elsewhere. While the simplest algorithms can only identify a few keywords, more recent developments are offering the possibility to extract knowledge graphs (See Appendix C).

A knowledge graph, also known as linked data, express knowledge in terms of nodes and edges between them. In the special case where nodes are concepts and edges represent causal relations, a knowledge graph is also a cognitive map. Thus, cognitive maps are a subset of knowledge graphs.

For illustrative purposes, Figure (7) shows a knowledge graph extracted from a medical text that illustrates the working principles of two alternative treatments, based on two different drugs acting on a gene and a protein, respectively. It is based on concepts (a disease, two drugs, a gene, a protein) linked to one another by causal relations and processes (treatment, inhibition). A cognitive map can be derived from this knowledge graph by associating processes to edges instead of diamond shapes.

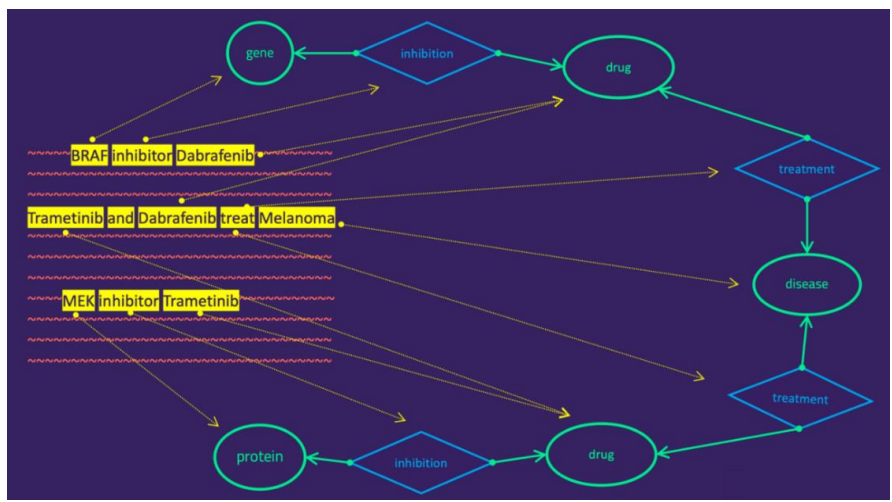

Figure 7. A knowledge graph extracted from a medical text. On the left, both *Trametinib* and *Dabrafenib* can be used to treat *Melanoma* but *Trametinib* inhibits protein MEK whereas *Dabrafenib* inhibits gene BRAF. By courtesy of Syed Irtaza Raza (2019), TypeDB. In order to derive a cognitive map, the processes in diamonds should be used to qualify causal relations

between concepts expressed in ovals. Both concepts and causal relations should be detailed by means of the words highlighted in yellow, on the left.

ML algorithms can be used to extract relevant graphs for decision problems where possibilities are known, but their probabilities are not (known unknowns), as well as for decision problems where doubts exist regarding the composition of the possibility set (unknown unknowns). In the first case, ML algorithms can procure the raw data to which the standard tools of *Q*-Analysis can be applied, whereas in the second case some work on defining the correct indicators is still waiting to be done.

As an instance where ML has been applied to contexts where "known unknowns" can exist we cite Rodrigues (2014), who employed ML in order to mine data from newspaper articles. By relating newspaper articles to one another along multiple dimensions a hypergraph ensued, whose structure could be investigated by means of *Q*-Analysis. In principle, further analyses could explore this hypergraph in order to highlight the structure of the consequences of certain decisions, for instance in terms of news diffusion.

The case of "unknown unknowns" is more complex. As a first step in this direction, we analysed the speeches held by *Lufthansa* CEO Carsten Spohr at annual general meetings (AGMs) before and after Covid-19, which has been a disruptive, unexpected and unthinkable event for airlines worldwide. In 2020, two indicators called *Density* and *Average Centrality* (see Appendix C) dropped by 43% and 53% compared to 2019. Since no comparable figure occurred in previous or subsequent years, we tentatively hypothesize that this drop was caused by the unexpected, unthinkable event. This analysis is at best preliminary, but it suggests that in principle it is possible to carry out massive evaluations of documents in order to detect the decision not to decide.

Furthermore, this possibility may open up interesting perspectives insofar as it concerns the availability of appropriate texts. Generally speaking, it is rather difficult to find texts where some decision-maker admits to have been in such a desperate situation that their cognitive map got disrupted. The 2020 Lufthansa AGM is probably an exception due to the fact that (*i*) the impact of Covid-19 was undeniable, (*ii*) it was clearly beyond the CEO's responsibility and, last but not least, (*iii*) the European Union had provided compensations that had largely offset the impact of the pandemic on airlines. Figure (8) illustrates a few key passages of Mr. Spohr's speech along with portions of the corresponding automatically generated cognitive map.

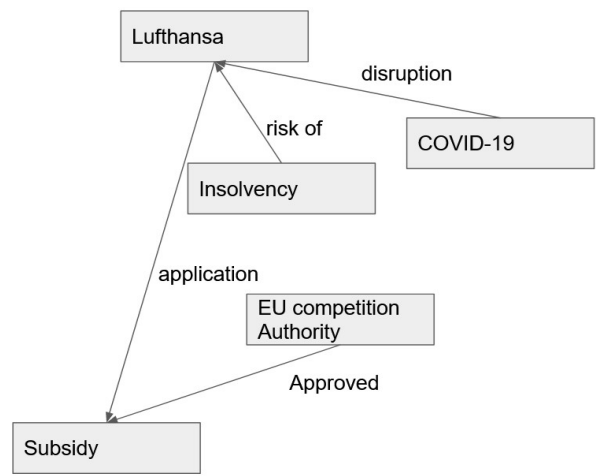

*However, our success story was abruptly interrupted at the beginning of the year by the outbreak of the Corona pandemic and its consequences.*

*Ladies and Gentlemen, Lufthansa's insolvency – also in the special form of protective shield proceedings – would lead to an almost complete loss of your share capital.*

*The competition authorities of the EU Commission have just today approved this package.*

Figure 8. Lufthansa 2020 AGM. A portion of the ML-generated cognitive map, along with key portions of the speech held by Lufthansa CEO Carsten Spohr.

However, ML can positively contribute to this problem by exploring the Internet looking for non-official sources such as blogs and social media where people eventually express themselves without the straitjackets imposed by the organizations they are members of. While this is still purely speculative for the time being, we are confident that the exploration of the Internet on a large scale could potentially yield substantial improvements to the data we are able to access.

One advantage of ML-based cognitive maps is that concepts and links are always extracted exactly in the same way, no matter which text is submitted or which researcher is doing the job. By contrast, when manual extraction of cognitive maps is carried out, uniformity has to be sought – and never fully achieved – by assigning the same text to several human coders and using only those concepts and causal relations that all coders identified (Sigismund Huff, 1990).

However, precisely uniformity constrains the quality of ML-based cognitive maps. Consider for instance the choice of a parameter such as entropy, which determines the coarseness of concept identification. If this parameter is set too loosely, concepts that are not appropriate for the text may be identified, whereas if it is too strict relevant concepts may be missed. By contrast, human researchers are capable to adapt to the text they are analysing, making human extraction more detailed, accurate and fine-grained.

We do not expect human-based extraction of cognitive maps to disappear altogether. In our opinion, one likely future is that human extraction will be used in order to calibrate the parameters of machine-based extraction.

**Conclusions**

For decades, non-probabilistic uncertainty has been the exotic province of critical economists and a handful of humanists and logicians who had little or no impact on applied science. However, since at least a decade this picture has started to change and, considering the sheer number of special issues of respected scientific journals being devoted to uncertainty, the trend is clearly accelerating. In our opinion, at least two forces are contributing to reverse the tide.

The first one, endogenous to Science, stems from realizing that the limitations of Probability Theory are due to its neglect of cognition. Probability Theory assumes that a set of possibilities is given, and that each possibility comes with a probability judgement. But what if decision makers have a hard time at assigning numerical values to these probabilities? What if they are unsure about the set of possibilities they are envisaging? Such questions cannot be faced unless one accepts to take cognition in serious consideration, making disciplines and research fields ranging from psychology to neural networks relevant for our understanding of uncertain reasoning. While in the 1970s a typical psychology article would attempt to fit decision into the straightjackets of Probability Theory, today the opposite is increasingly taking place, with cognitive sciences capable of addressing uncertain reasoning while ignoring the strictures and assumptions of Probability Theory.

The second force moving Science in this direction is, in our opinion, the sheer fact that the technological, environmental and political stage has become uncertain in ways that used to be unthinkable just a few years ago. Nowadays, with small-sized nuclear weapons having been adopted by all major armies, the prospect of a nuclear war has become a feasible nightmare of an endless series of targeted annihilations. Epidemics, once relegated to the Dark Ages, appear to be perfectly possible. The financial crises of the XIX century have resurrected in the XXI, and climate change is posing terribly deep and disturbing questions to which no answer is in sight. In such an environment, the existence of non-probabilistic uncertainty is obvious for scientists and laypersons alike.

We did not contribute any fundamentally novel concept or tool to this trend, but rather assembled previously disconnected pieces of knowledge in a novel way, pointing to the possibilities offered by machine learning. We proposed a research path that moved from adding cognitive insights to argumentations that used to be expressed in purely logical terms, to subsequently describe non-probabilistic uncertainty in terms of procedures and sequences of events that make it emerge rather than static definitions, onwards to mathematical and computational tools to observe it, and – finally – machine learning as a means to carry out observations on a large scale. We firmly believe that this path has a rich potential for meaningful and useful discoveries, but we are aware of the difficulties posed by the fact that both individuals and organizations are often unwilling to disclose their helplessness in front of missing information or, even worse, contradictory and disruptive information that destroys their cognitive maps. Great care must be exercised in selecting proper sources, which may not be limited to third parties as in the case of a dedicated consultant for BioTech companies but may include unofficial or semi-official sources such as web blogs. In this respect, just like on many other issues surrounding Big Data and the Internet, manipulation of information looms as a substantial danger to the effectiveness of ML-based methods, potentially able to destroy the reliability of tools that have been developed under the assumption that information may be possibly difficult to find, but reliable.

**Appendix A: *Q*-Analysis**

This appendix illustrates the basics of *Q*-Analysis (Atkin, 1974; Johnson, 1990; Fioretti, 2023). Please consult these references for further details.

A *simplex* is the convex hull of a set of $(n + 1)$ independent points in some Euclidean space of dimension *n* or higher. These points are its *vertices*. A 0-dimensional simplex is a point, a 1-dimensional simplex is a segment, a 2-dimensional simplex is a triangle. Henceforth, higher-dimensional simplices will not be considered.

The convex hull of any non-empty subset of the points that define a simplex is called a *face* of the simplex. In particular, 0-dimensional faces are the vertices of a simplex, 1-dimensional faces are segments that connect vertices. Two simplices are connected if they have a common face. A set of (at least) pairwise connected simplices is a *simplicial family*, which corresponds to a *hypergraph* where simplices are the sets of nodes that an edge can connect. In the special case where each face of each simplex also belongs to the simplicial family, this is called a *simplicial complex*. Although *Q*-Analysis was first conceived for simplicial complexes, it actually applies to simplicial families of any sort.

Henceforth, a cognitive map is graphically represented as a simplicial family. Evoked Alternatives are simplices whose vertices are the Perceived Consequences to which each Evoked Alternative is connected. Figure (A1) illustrates the simplicial families that correspond to cases (b) and (c) in Figure (2).

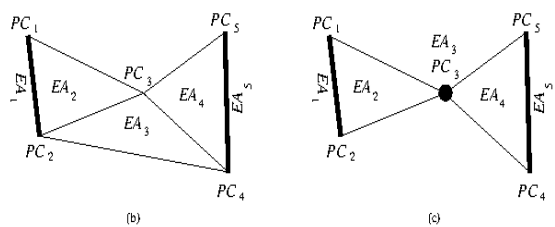

Figure A1. Left, the simplicial family corresponding to case (b) of Figure (2). Right, the simplicial family corresponding to case (c) of Figure (2). Segments representing 1-dimensional simplices are thicker than segments representing the faces of 2-dimensional simplices (triangles).

If the connections between categories of actions and categories of results are all one-to-one as in case (a) of Figure (2), then the simplices are isolated points so no connected simplicial family exists. In this case, complexity is zero. By contrast, in case (b) and (c) the simplicial families illustrated in Figure (A1) yield a complexity greater than zero.

Two simplices are connected if they have at least one common vertex. Two simplices that have no common vertex may nonetheless be connected by a chain of simplices having common vertices with one another. Let us say that simplices $EA_i$ and $EA_j$ are $q-connected$ if there exists a chain of simplices $EA_u, EA_v, \ldots EA_w$ such that $q = min\{l_{i,u}, l_{u,v}, \ldots l_{w,j}\} \geq 0$, where $l_{x,y}$ is the dimension of the common face between $EA_x$ and $EA_y$. In particular, two contiguous simplices are connected at level $q$ if they have a common face of dimension $q$.

Let us consider the common faces between simplices and let us focus on the face of largest dimension. Let $Q$ denote the dimension of this face. It is necessarily $Q \leq n - 1$, where $Q = n - 1$ means that there are at least two overlapping simplices that include all possible vertices.

Let us partition the set of simplices that compose the simplicial family according to their connection level $q$. In general, for $\forall q$ there exist several classes of simplices such that the simplices belonging to a class are connected at level $q$. Let us introduce a *structure vector* **s** whose $q$-th component $s_q$ denotes the number of disjoint classes of simplices that are connected at level $q$. Since $q = 0, 1, \ldots Q$, vector **s** has $Q + 1$ rows.

Let us define *Complexity* (Fioretti, 2001) as:

$$C = \begin{cases} 0 & \text{if all links are one to one} \\ \sum_{q=0}^{Q} \frac{q+1}{s_q} & \text{otherwise} \end{cases}$$

where the sum extends only to all terms such that $s_q \neq 0$. The complexity of two or more disconnected simplicial families is the sum of their complexities.

This expression takes account of two opposite effects. On the one hand, its numerator increases with the number of connections between Evoked Alternatives and Perceived Consequences. Thus, it simply measures the extent to which novel connections confuse the cognitive map. On the other hand, the denominator makes complexity decrease to the extent that cross-connections are clustered in distinct groups.

In case (b) of Figures (2) and (A1) there exists one single class of simplices connected at level $q = 0$ and one single class of simplices connected at level $q = 1$, hence $C = \frac{(0+1)}{1} + \frac{(1+1)}{1} = 3$. By contrast, in case (c) there exists one class of simplices connected at level $q = 0$ but two classes of simplices connected at level $q = 1$, hence $C = \frac{(0+1)}{1} + \frac{(1+1)}{2} = 2$.

**Appendix B: The Series of Cognitive Maps**

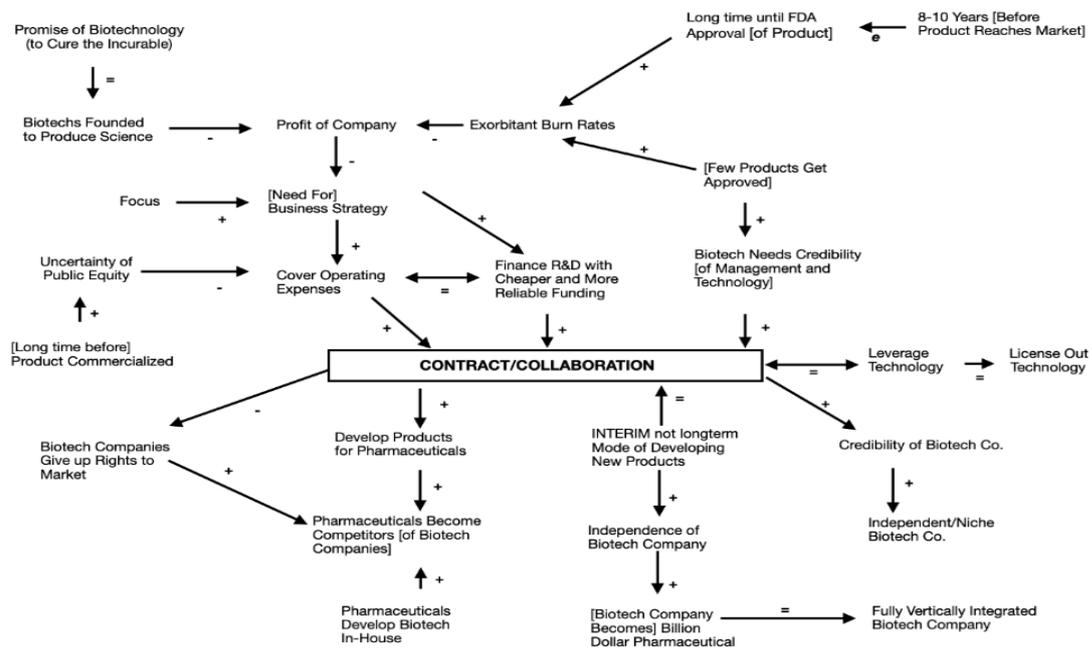

Figure B1. Biotech companies' cognitive map, 1986 (James, 1996).

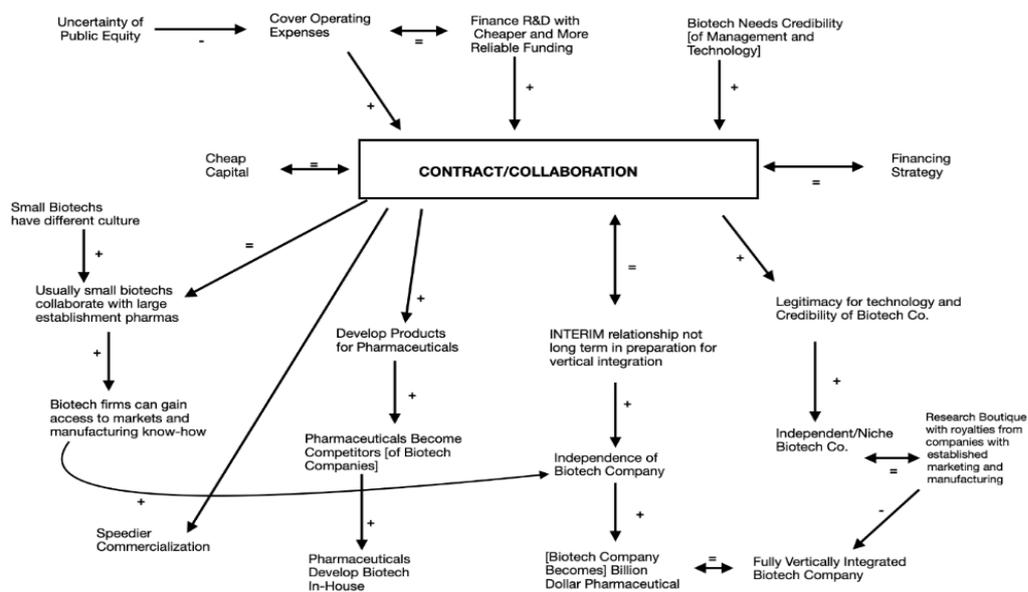

Figure B2. Biotech companies' cognitive map, 1987 (James, 1996).

Figure B3. Biotech companies' cognitive map, 1988 (James, 1996).

Figure B4. Biotech companies' cognitive map, 1989 (James, 1996).

Figure B5. Biotech companies' cognitive map, 1990 (James, 1996).

Figure B6. Biotech companies' cognitive map, 1991 (James, 1996).

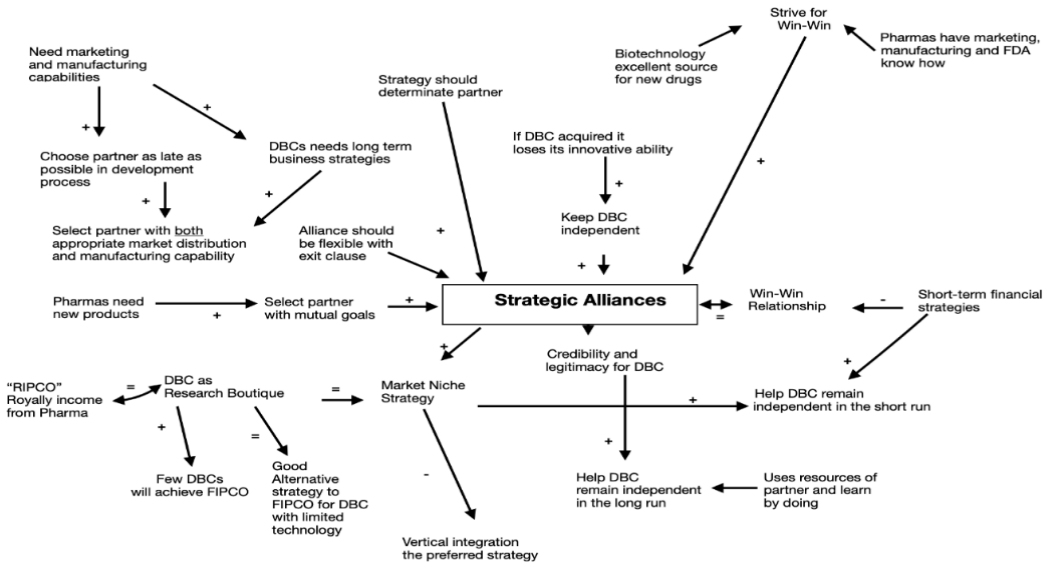

Figure B7. Biotech companies' cognitive map, 1992 (James, 1996).

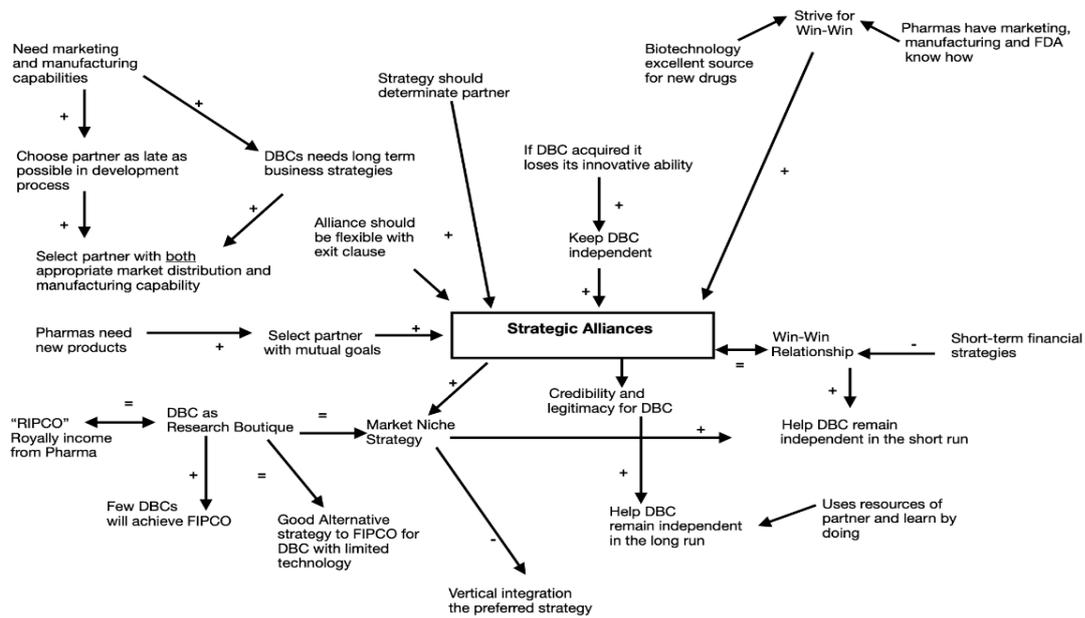

Figure B8. Biotech companies' cognitive map, 1993 (James, 1996).

**Appendix C: Machine Learning**

The last 10 years witnessed a rapid development in Natural Language Processing (NLP), which uses Machine Learning (ML) algorithms to allow computers to process and "understand" text semantics (Ruder, 2022). As a result, ML algorithms have surpassed human performance on a number of NLP tasks such as aspects of speech tagging or question and answering (Stanford NLP Group, 2022).

These developments build on previous advances in knowledge representation, which historically had been one of the key purposes of "good old fashioned" Artificial Intelligence (AI) with the aim of building general or domain-specific cognitive maps.

Cognitive maps can be built from texts to represent knowledge in terms of relations between entities. Once knowledge has been represented as a cognitive map, the graph can be traversed to draw relevant conclusions such as, e.g., whether *What'sApp* belongs to *Facebook* or *Joe Biden* is the *President of the United States*. However, we focus on the graph structure instead of its traversal. Since 2019 there has been a revival of the attempt to combine statistical models that extract the meaning of texts with automatic knowledge representation in the form of graph neural networks (Zhang *et al*., 2020).

We implement machine generation of cognitive maps from text through a pipeline composed by seven main steps, as illustrated in Figure C1. The first step is pre-processing. The second step is concept recognition, which focuses on identifying the concepts which will be connected in the cognitive map. The third and last step is relational identification, where the linkages are determined.

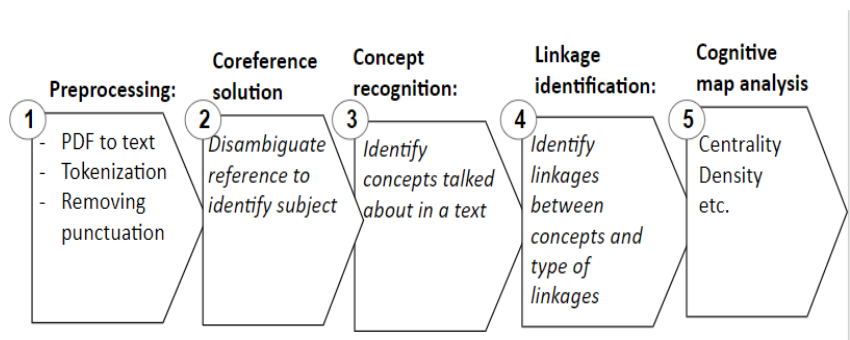

Figure C1. The pipeline for automatically generating cognitive maps from texts.

*Step 1: pre-processing.*

The original source is first converted into a raw text file, which is then split into individual sentences and tokenized. Tokenization involves splitting up the entities of a sentence, words and punctuation into individual components. In our case, these components are embedded as vectors and punctuation is removed. Using word embeddings is a typical process in natural language processing, where each word is represented as, e.g., a 124-dimensional vector in a latent space (Mikolov *et al.*, 2013). The vector represents the meaning of the word. The advantage of such word embedding is that words such as *Cappuccino* and *Latte Macchiato*, for example, are related in their meaning (cosine distance between their embedding vectors), when such relatedness cannot be recognized by a computer form their surface form (i.e., the letters of the words).

*Step 2: Coreference solution.*

Next the problem of co-reference is addressed, which is the task of finding all the expressions that refer to the same entity in the text (Clark & Manning, 2016). For instance, pronouns must be substituted by the nouns they refer to: "X is a public company. It made a loss in 2021" becomes "X is a public company. X made a loss in 2021". This task is achieved by a neural network which has been trained on a dataset where coreference has been resolved by human annotators.

*Step 3: Concept recognition.*

In order to recognize concepts, a machine must be trained on a meaningful network of concepts. In this step the concepts talked about in the text are matched to concepts covered on Wikipedia. We take a neural network pretrained on 6.4 million concepts entailed in Wikipedia (Brank, Leban & Grobelnik, 2017). Whilst not every concept a CEO could talk about might be entailed in Wikipedia, the concepts described on Wikipedia are generally regarded as a reasonable baseline. The ground truth on which the model is trained is again a dataset of texts where human annotators have classified which concept on Wikipedia they are related to.

*Step 4: Linkage extraction*

Once the concepts have been identified, the linkages between them must be identified. The algorithm needs to identify what concept is related to another concept according to the text and how they are related. For this purpose, we take a pre-trained model. The BERT transformer is currently the model with the highest accuracy (Devlin *et al.*, 2019) for this task.

*Step 5: Cognitive map analysis*

Having established the cognitive map, we need to be able to systematically measure how it changes over time or between companies in order to detect the effects of non- probabilistic uncertainty. One way of doing so is to measure the number of concepts and linkages in a text. However, the number of concepts and linkages dependent on the length of the text. A ratio of the number of links per concept can normalize for differences in the length of the text.

If cognitive maps are disrupted, the connectedness of concepts should change. Such change can be measured in two metrics. The first one is the cognitive map density, which is the ratio of the number of linkages in the network to the maximum number of possible linkages:

$$Density = \frac{|L|}{|L|_{max}}$$

where $|L|$ denotes the number of linkages. The maximum number of linkages for a directed graph is defined as follows:

$$|L|_{max} = \frac{|C| * (|C| - 1)}{2}$$

where $C$ is the set of concepts and $|C|$ denotes the number of concepts.

Another way in which the evolution of cognitive maps can be measured over time is in terms of closeness centrality, which is the average distance of a concept to all other concepts. It is measured as the number of linkages one needs to travel to reach any other concept:

$$\overline{U} = \frac{|C| - 1}{\sum_{C_i} \sum_{C_j \in \{C - C_j\}} dist(C_i, C_j)}$$